\documentclass[10pt,twocolumn,letterpaper]{article}

\usepackage{cvpr}              

\usepackage[accsupp]{axessibility}
\definecolor{cvprblue}{rgb}{0.21,0.49,0.74}
\usepackage[pagebackref,breaklinks,colorlinks,allcolors=cvprblue]{hyperref}

\def\paperID{43} 
\def\confName{CVPR}
\def\confYear{2025}

\title{Frontier Vision-Language Models: Architectural Evolution, Benchmarks, Applications, and Challenges}

\author{\begin{minipage}{\textwidth}
    \centering
    \begin{tabular}{c@{\hskip 20pt}c@{\hskip 20pt}c}
      \textbf{Zongxia Li}$^{*}$ & \textbf{Xiyang Wu}$^{*}$ & \textbf{Hongyang Du} \\
    \end{tabular}
  \end{minipage} \\[7pt]
  \begin{minipage}{\textwidth}
    \centering
    \begin{tabular}{c@{\hskip 20pt}c@{\hskip 20pt}c}  
      \textbf{Fuxiao Liu} & \textbf{Huy Nghiem} & \textbf{Guangyao Shi} \\
    \end{tabular}
  \end{minipage} \\[10pt]
  \begin{minipage}{\textwidth}
    \centering
    \texttt{\{zli12321, wuxiyang, hdu1, fl3es, nghiemh\}@umd.edu\\shig@usc.edu}
  \end{minipage} \\[10pt]
  \\[5pt]
  {\small\textcolor{magenta}{\url{https://github.com/zli12321/Vision-Language-Models-Overview.git}}}
}

\newcommand{\abr}[1]{\textsc{#1}}
\usepackage{comment}
\usepackage{graphicx}
\usepackage{color}
\begin{document}
\maketitle

\begin{abstract}
Large vision-language models (\abr{vlm}s) have evolved rapidly from contrastive image-text encoders and adapter-based assistants into natively multimodal foundation models that support long-context reasoning, agentic workflows, and, increasingly, unified perception-generation-action loops. Since 2025, the frontier has consolidated around a small number of large pretraining families, including GPT, Gemini, Claude, Grok, Qwen, Gemma, DeepSeek, Kimi, and MiniMax, while evaluation has shifted from short-form visual question answering toward spatial, temporal, embodied, and calibration-sensitive benchmarks.
In this survey, we provide an updated overview of \abr{vlm}s through 2026 with an emphasis on three developments.
First, we trace the architectural evolution from contrastive or bridged two-tower models to \abr{llm}-backbone adapter models, native multimodal-input models, omni-modal unified input/output models, and emerging world-action models.
Second, we summarize how benchmarks expanded from classical OCR, VQA, and chart understanding toward long-video reasoning, embodied evaluation, reward-model judging, and domain-specific decision making.
Third, we review the growing role of post-training, including supervised fine-tuning, preference optimization, and reinforcement learning methods such as GRPO-style multimodal alignment.
Rather than exhaustively listing every release, we focus on representative frontier families, influential benchmarks, and the major open challenges that remain in hallucination, safety, efficiency, data quality, and multimodal alignment.
\end{abstract}
\label{abstract}

\section{Introduction}
Pretrained large language models (\abr{llm}s), such as LLaMA~\cite{touvron2023llama2openfoundation} and GPT-4~\cite{openai2024gpt4technicalreport}, have achieved remarkable success across a wide range of \abr{nlp} tasks~\cite{makridakis2023large, mohammad2023large}. As these systems continue to scale~\cite{nayab2024concise}, they face two persistent constraints: the finite supply of high-quality text data~\cite{villalobosposition, li2024beyond} and the inherent limitation of single-modality models for tasks that require joint visual, spatial, temporal, and linguistic reasoning~\cite{goodwin2018multimodality, hong2024only}.
These pressures motivate the development of large vision-language models (\abr{vlm}s), which combine visual inputs such as images and videos with text to produce richer representations of objects, scenes, events, and abstract concepts~\cite{bordes2024introduction, hartsock2024vision}. \abr{vlm}s already power visual question answering, web and GUI agents, document understanding, robotics, and autonomous driving~\cite{agrawal2016vqavisualquestionanswering, tian2024drivevlmconvergenceautonomousdriving}. At the same time, they introduce failure modes that are weaker in text-only systems, including visual hallucination, brittle grounding, and shallow multimodal reasoning~\cite{guan2024hallusionbench, liu2024surveyhallucinationlargevisionlanguage}.
The field changed especially quickly in 2025--2026. Frontier systems are no longer isolated point releases, but fast-moving multimodal families from a small number of labs: GPT~\cite{openai2026gpt56}, Gemini~\cite{google2026gemini36flash}, Claude~\cite{anthropic2026claudeopus46}, Grok~\cite{xai2026grok45}, Qwen~\cite{qwen2026qwen38}, Gemma~\cite{google2026gemma4}, DeepSeek~\cite{wu2024deepseekvl2mixtureofexpertsvisionlanguagemodels}, Kimi~\cite{kimi2026k3}, and MiniMax~\cite{minimax2026m3}. Across these families, the center of gravity has shifted from attaching a vision encoder to a text model toward pretraining a single multimodal trunk with longer context, stronger tool use, and better support for agentic workflows.
In this paper, we provide a critical examination of modern \abr{vlm}s with three emphases: the architectural evolution of frontier systems, the expansion of evaluation beyond classical VQA-style tests, and the growing role of post-training and reinforcement learning in multimodal reasoning. Rather than attempting an exhaustive live index of every release, we focus on representative frontier families and benchmark directions that clarify how the field is changing.

\section{State-of-the-Art \abr{vlm}s}
\begin{figure*}[!t]
    \centering
    \includegraphics[width=0.97\textwidth]{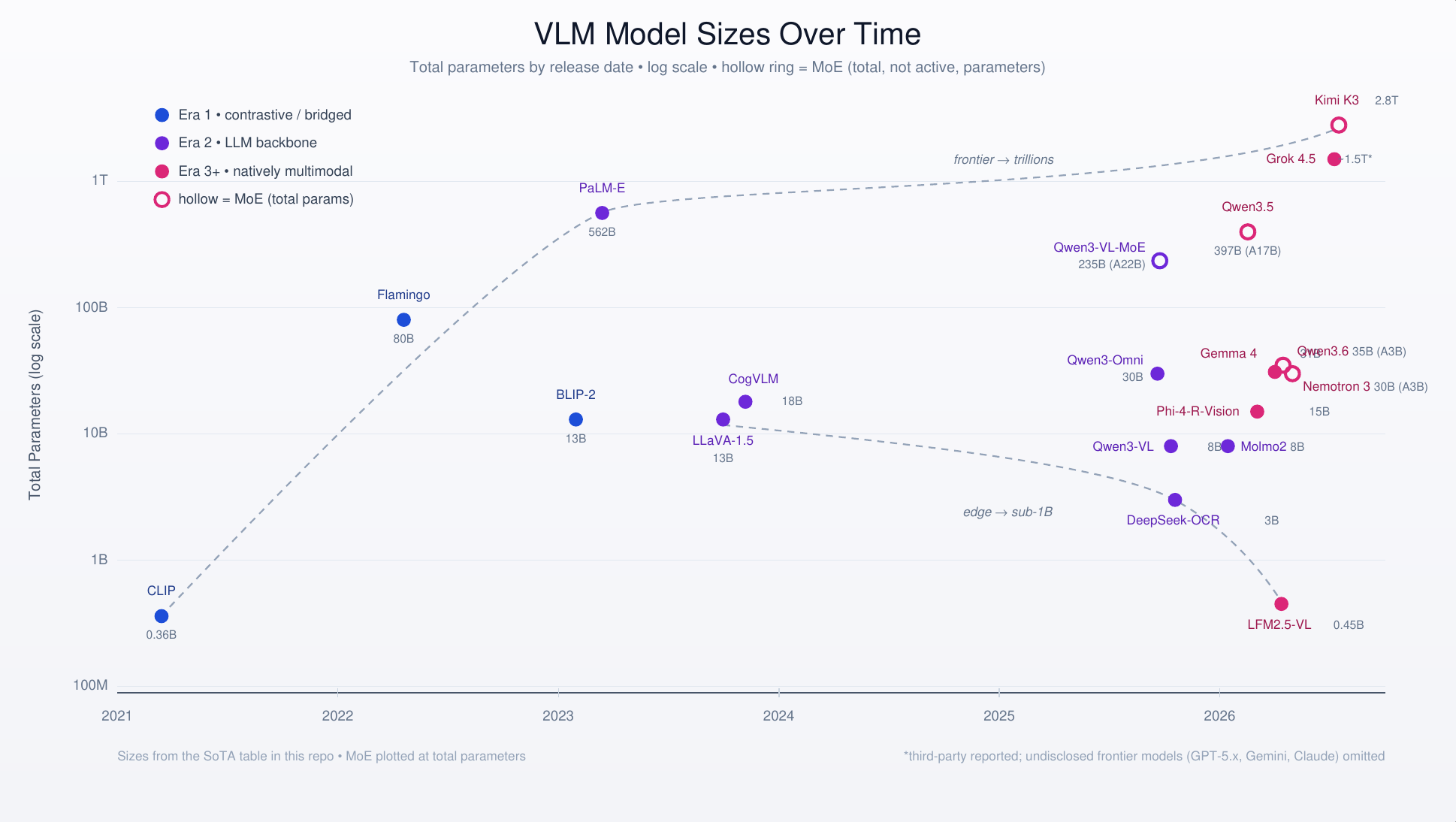}
    \caption{Selective timeline of representative \abr{vlm} families. Hollow markers denote mixture-of-experts models plotted by total rather than active parameters. Frontier proprietary families with undisclosed parameter counts, including GPT, Gemini, and Claude, are discussed in the text and Table~\ref{table:VLMlist} but cannot be placed exactly on the curve.}
    \label{fig:vlm_timeline}
\end{figure*}

Recent \abr{vlm} progress is best understood as the evolution of a small number of pretraining families rather than a flat list of independent models. Figure~\ref{fig:vlm_timeline} highlights how public model scale grew from sub-billion contrastive systems to frontier multimodal families, while Table~\ref{table:VLMlist} summarizes representative brand-name releases that define the current landscape.

\begin{table*}[!t]
\centering
\footnotesize
\setlength{\tabcolsep}{4pt}
\renewcommand{\arraystretch}{1.12}
\begin{tabular}{p{0.12\textwidth}p{0.28\textwidth}p{0.08\textwidth}p{0.14\textwidth}p{0.24\textwidth}}
\toprule
\textbf{Family} & \textbf{Representative releases} & \textbf{Era} & \textbf{Public scale} & \textbf{Snapshot} \\
\midrule
OpenAI & GPT-4V~\cite{yang2023dawn}, GPT-5.6 Sol~\cite{openai2026gpt56} & 2 $\rightarrow$ 3a & Undisclosed & Closed frontier family for multimodal reasoning, coding, and computer use \\

Google DeepMind & Gemini~\cite{anil2023gemini}, Gemini 3.6 Flash~\cite{google2026gemini36flash} & 2 $\rightarrow$ 3a & Undisclosed & Workhorse multimodal family with 1M-context agentic reasoning \\

Anthropic & Claude 3~\cite{claude2024}, Claude Opus 4.6~\cite{anthropic2026claudeopus46} & 2 $\rightarrow$ 3a & Undisclosed & Long-context multimodal family focused on code and agentic workflows \\

xAI & Grok 4.5~\cite{xai2026grok45, xai2026grok45docs} & 3a & Reported $\sim$1.5T & Vision-capable coding and agentic model with 500K context \\

Alibaba Qwen & Qwen-VL~\cite{bai2023qwenvlversatilevisionlanguagemodel}, Qwen2.5-VL~\cite{bai2025qwen25vltechnicalreport}, Qwen3.6-35B-A3B~\cite{qwen2026qwen36}, Qwen3.8-Max~\cite{qwen2026qwen38} & 2 $\rightarrow$ 3a/3b & 2.4T / 95B active & Family now spans compact open checkpoints and a 2.4T multimodal flagship \\

Google Gemma & Gemma 4~\cite{google2026gemma4} & 3a & 31B dense; 26B MoE (3.8B active) & Compact open multimodal family built from the same research as Gemini 3 \\

DeepSeek & DeepSeek-VL2~\cite{wu2024deepseekvl2mixtureofexpertsvisionlanguagemodels}, DeepSeek-OCR-2~\cite{wei2026deepseekocr2} & 2 & 4.5B active (VL2) & Open multimodal family with strong OCR, document, and vision-language reasoning \\

Moonshot Kimi & Kimi K3~\cite{kimi2026k3} & 3a & 2.8T / 104B active & Open 3T-class native multimodal flagship with 1M context \\

MiniMax & MiniMax M3~\cite{minimax2026m3} & 3a & 428B / 23B active & Native multimodal 1M-context model for coding, video, and computer use \\
\bottomrule
\end{tabular}
\caption{Selective frontier \abr{vlm} families through 2026. We emphasize major pretraining families from large labs rather than attempting to list every derived checkpoint or task-specific open model.}
\label{table:VLMlist}
\end{table*}

Two shifts stand out from Table~\ref{table:VLMlist}. First, the dominant units of progress are now families, each with multiple reasoning modes, context tiers, and deployment targets. Second, architectural progress is no longer well described by the simple story of "add a vision encoder to an \abr{llm}." The frontier moved from contrastive or bridged two-tower systems such as CLIP~\cite{radford2021learningtransferablevisualmodels} and Flamingo~\cite{alayrac2022flamingo}, to \abr{llm}-backbone adapter models, and then to native multimodal pretraining where image, video, audio, and text share a single trunk.

\begin{figure*}[!thbp]
    \centering
    \includegraphics[width=0.97\textwidth]{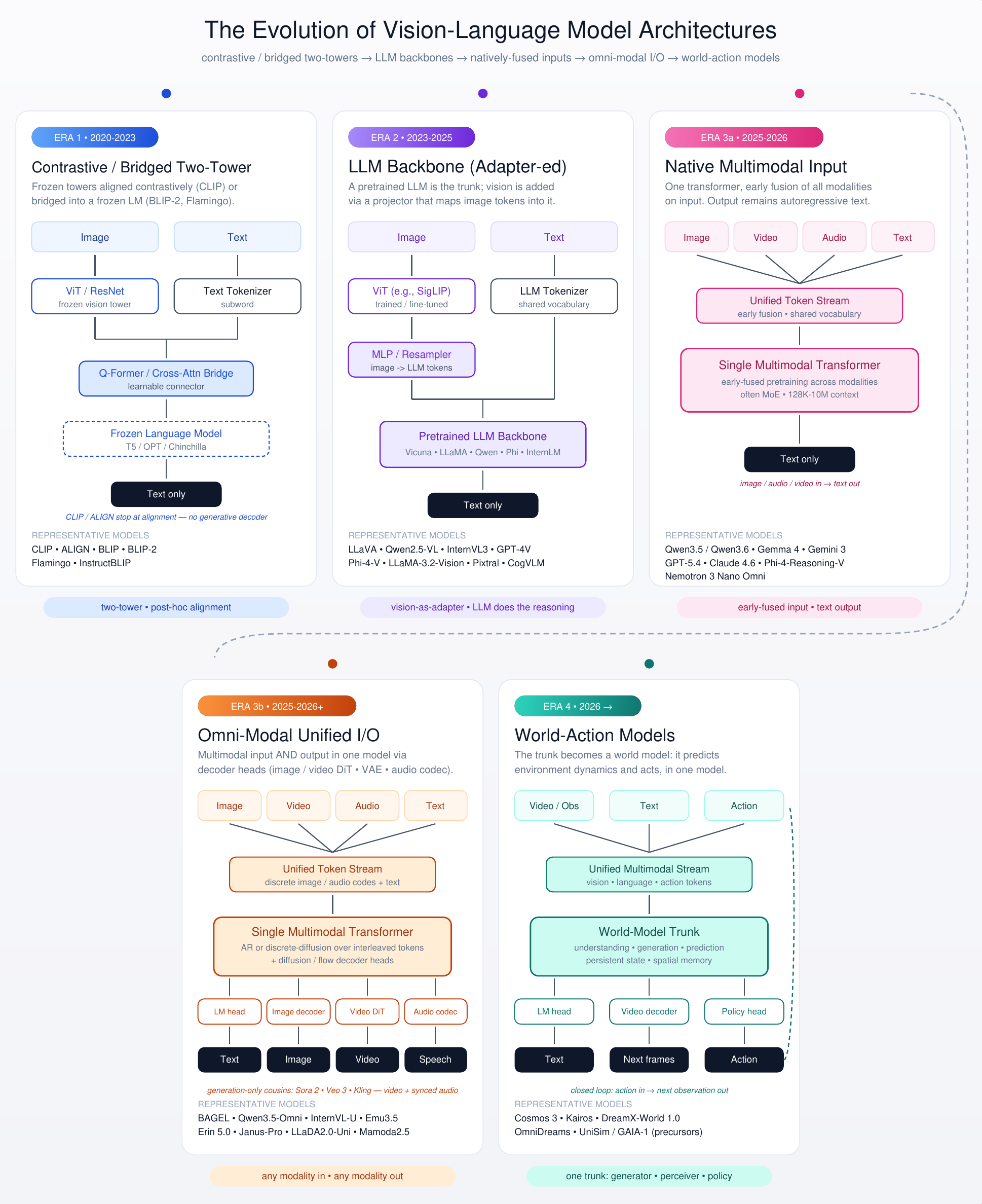}
    \caption{Architectural evolution of modern \abr{vlm}s. Era 1 aligns separate vision and language towers contrastively or through a learned bridge. Era 2 makes a pretrained \abr{llm} the trunk and injects visual tokens through a projector or resampler. Era 3a early-fuses multimodal inputs into a single transformer but still emits text. Era 3b extends the same trunk with image, video, and audio decoder heads for unified input/output generation. Era 4 adds action and persistent state, yielding emerging world-action models that act as generator, perceiver, and policy in one network.}
    \label{fig:vlmarc}
\end{figure*}

We use this architectural view throughout the rest of the paper. Section~\ref{sec:architecture} expands the building blocks behind each era. Section~\ref{sec:benchmarks} examines how evaluation evolved from short-form answer matching toward spatial, temporal, embodied, and calibration-aware assessment. Section~\ref{sec:application} then connects these capabilities to computer-use agents, document intelligence, media generation, and embodied systems. Finally, the later sections discuss why post-training, alignment, safety, and data quality have become central bottlenecks as frontier \abr{vlm} families grow more agentic and multimodal.

\label{introduction}

\section{Building Blocks and Training Methods}
\label{sec:architecture}

The evolution of \abr{vlm}s is no longer captured by a single transition from "train from scratch" to "attach vision to an \abr{llm}." As Figure~\ref{fig:vlmarc} and Table~\ref{table:VLMlist} show, the field now spans several architectural eras: contrastive or bridged two-tower systems, projector-based \abr{llm}-backbone models, native multimodal-input models, unified input/output models, and emerging world-action models.
Across these eras, the same ingredients still recur: a visual tokenizer or encoder, a language-side tokenizer and decoder, alignment modules or shared token spaces, and increasingly sophisticated post-training. What changed is where modality fusion happens, whether outputs remain text-only, and whether the model is expected merely to describe the world or also to generate and act within it.
In this section, we summarize the most foundational architectural components of \abr{vlm}s and then explain how frontier systems reorganize them across eras. We conclude with the recent post-training shift from supervised instruction tuning toward preference optimization and reinforcement learning for multimodal reasoning.

\subsection{Common Architecture Components}

\noindent \textbf{Vision Encoder} plays a crucial role in projecting visual components into embedding features that align with embeddings from large language models (LLMs) for tasks such as text or image generation~\cite{multimodal-autoregressive}. 
It is trained to extract rich visual features from image or video data, enabling integration with language representations~\cite{maaz2024videogptintegratingimagevideo, zhao2024videoprismfoundationalvisualencoder}.

Specifically, vision encoders used in many VLMs~\cite{liu2023visualinstructiontuning, wang2024qwen2vlenhancingvisionlanguagemodels, dai2024nvlmopenfrontierclassmultimodal, chen2024internvlscalingvisionfoundation}, are pretrained on large-scale multimodal or image data: These encoders are jointly trained on image-text pairs, allowing them to capture visual and language relationships effectively. Notable examples include CLIP~\cite{radford2021learningtransferablevisualmodels}, which aligns images and text embeddings via contrastive learning, and BLIP~\cite{li2022blipbootstrappinglanguageimagepretraining}, which leverages bootstrapped pretraining for robust language-image alignment.
Pretrained on large scale ImageNet~\cite{ImageNet} or Similar Datasets: These encoders are trained on vast amounts of labeled visual data or through self-supervised training~\cite{pathak2016contextencodersfeaturelearning}, enabling them to capture domain-specific visual features. While initially unimodal, these encoders, such as ResNet~\cite{he2015deepresiduallearningimage} or Vision Transformers (ViTs)~\cite{dosovitskiy2021imageworth16x16words}, can be adapted for multimodal tasks. 
They excel at extracting meaningful object-level features and serve as a solid foundation for vision-language models.
Many SoTA VLMs, such as Qwen2-VL~\cite{wang2024qwen2vlenhancingvisionlanguagemodels} and LLaVA~\cite{liu2023visualinstructiontuning}, commonly incorporate pretrained vision encoders. 
These encoders not only provide robust and meaningful visual representations but are also highly effective for transfer learning~\cite{zamir2018taskonomydisentanglingtasktransfer}. 
They outperform randomly initialized encoders~\cite{holmberg2020self} by leveraging learned vision knowledge from their training domains.

\noindent \textbf{Text Encoder} projects tokenized text sequences into an embedding space, similar to how vision encoders process images.
Models such as CLIP~\cite{radford2021learningtransferablevisualmodels}, BLIP~\cite{li2022blipbootstrappinglanguageimagepretraining}, and ALIGN~\cite{jia2021scalingvisualvisionlanguagerepresentation} use both an image encoder and a text encoder. 
These models use contrastive learning to align image and text embeddings in a shared latent space, effectively capturing cross-modal relationships.
However, newer models, such as LLaVA~\cite{liu2023visualinstructiontuning}, often do not include a dedicated text encoder. 
Instead, they rely on large language models (LLMs) (e.g., LLaMA~\cite{touvron2023llama2openfoundation}, Vicuna~\cite{vicuna2023}) for text understanding, integrating visual inputs through projection layers or cross-attention mechanisms~\cite{lin2021catcrossattentionvision}. 
This shift shows a growing trend of using the capabilities of LLMs over vision components for more versatile and advanced multimodal reasoning and generation tasks.

\noindent \textbf{Text Decoder} leverages \abr{llm}s as the primary text generator, using visual encoders to project image features~\cite{kim2021viltvisionandlanguagetransformerconvolution}. 
GPT-4V~\cite{openai2024gpt4technicalreport}, Flamingo~\cite{alayrac2022flamingo}, and Kosmos-2~\cite{peng2023kosmos2groundingmultimodallarge} use this approach. 
These models typically use a minimal visual projection mechanism, allowing the powerful language decoder to generate contextually rich outputs. 
ViLBERT and VisualBERT~\cite{lu2019vilbertpretrainingtaskagnosticvisiolinguistic, li2019visualbertsimpleperformantbaseline} provide the foundation to decoder architectures for multimodal pretraining.
Training \abr{vlm}s from scratch typically requires a separate text decoder, whereas using \abr{llm}s as the backbone often uses the original decoders from the \abr{llm}. (Figure~\ref{fig:vlmarc}).

%

\noindent \textbf{Cross-Attention Mechanisms} enable visual-text interactions by allowing tokens from one modality (vision) to influence tokens from the other modality (text)~\cite{lin2021catcrossattentionvision}. These layers compute attention scores across modalities, but not all models use them. ViLBERT~\cite{lu2019vilbertpretrainingtaskagnosticvisiolinguistic} and Flamingo~\cite{alayrac2022flamingo} employ cross-attention, while CLIP~\cite{radford2021learningtransferablevisualmodels} does not.

\subsection{Building Blocks of Training From Scratch}
Training a \abr{vlm} from scratch typically uses distinct training objectives and methodologies compared to using an \abr{llm} as the backbone.
Self-Supervised Learning (\abr{ssl}) pre-trains without needing human labeled data to scale up pretraining~\cite{he2020momentumcontrastunsupervisedvisual}. 
Variants of \abr{ssl} techniques include masked image modeling~\cite{he2021maskedautoencodersscalablevision}, contrastive learning~\cite{oord2019representationlearningcontrastivepredictive}, and image transformation prediction~\cite{misra2019selfsupervisedlearningpretextinvariantrepresentations}. 
In this section, we delve into contrastive learning, a common pre-training process to scale up \abr{vlm} training from scratch.

\noindent \textbf{Contrastive Learning} employs separate encoders for visual and textual inputs, mapping them into a shared embedding space. The visual encoder extracts features using 
convolutional neural networks (\abr{cnn})~\cite{oshea2015introductionconvolutionalneuralnetworks} or vision transformers (ViTs)~\cite{dosovitskiy2021imageworth16x16words}. 
The text encoder processes textual inputs into embeddings.
Contrastive learning aligns related image-text pairs by minimizing the distance between their visual and text embeddings in the shared space,
while maximizing the distance between embeddings of unrelated pairs.
Pioneering models like CLIP~\cite{radford2021learningtransferablevisualmodels}, BLIP~\cite{li2022blipbootstrappinglanguageimagepretraining}, and ALIGN~\cite{jia2021scalingvisualvisionlanguagerepresentation} leverage this approach, pre-training on large-scale image-text datasets to develop robust, transferable representations for downstream tasks.




\subsection{Building Blocks of Using \abr{llm}s as Backbone}
\noindent \textbf{Large Language Models} serve as the text generation component that processes encoded visual and textual inputs to produce text outputs autoregressively~\cite{brown2020languagemodelsfewshotlearners, touvron2023llama2openfoundation, openai2024gpt4technicalreport} for \abr{vlm}s.
In the context of \abr{vlm}s, \abr{llm}s include their original text decoders.
In this section, we list two common ways to align visual and pre-trained \abr{llm} text features.

\noindent \textbf{Projector} maps visual features extracted by the vision encoder into a shared embedding space aligned with the text embeddings from the \abr{llm}. 
It typically consists of multi-layer perceptron (\abr{mlp}) layers~\cite{MURTAGH1991183}, which transform high-dimensional visual representations into compact embedding tokens compatible with the textual modality.
The projector can be trained jointly with the rest of the model to optimize cross-modal objectives or freezing certain parts of the model, such as the \abr{llm}, to preserve pre-trained knowledge.
Projectors are the defining connector of Era 2 systems. Representative examples include LLaVA~\cite{liu2023visualinstructiontuning}, Qwen2-VL~\cite{wang2024qwen2vlenhancingvisionlanguagemodels}, DeepSeek-VL2~\cite{wu2024deepseekvl2mixtureofexpertsvisionlanguagemodels}, and Pixtral~\cite{agrawal2024pixtral12b}. In newer native multimodal families, the explicit projector becomes smaller, more distributed, or disappears altogether because all modalities are pretrained into a single trunk.

\noindent \textbf{Joint Training} is an end-to-end approach that updates weights of all components of the model in parallel without freezing any weights, including the \abr{llm} and projector layers. 
This approach has been used in models such as Flamingo~\cite{alayrac2022flamingo}.

\noindent \textbf{Freeze Training Stages} involves selectively freezing model components during training, preserving pre-trained knowledge while adapting to new tasks~\cite{howard2018universallanguagemodelfinetuning}. 
Common strategies include freezing pre-trained vision encoders while fine-tuning projector layers, and implementing gradual unfreezing of components~\cite{peters2019tunetuneadaptingpretrained} or freezing \abr{llm} layers while only updating vision encoder weights~\cite{tsimpoukelli2021multimodalfewshotlearningfrozen}.

\subsection{Newer Architectures}
\label{sec:visual_as_tokens}
Recent frontier work is less about inventing new standalone components than about changing where fusion happens and what the multimodal trunk is expected to output.

\noindent \textbf{Era 2: projector-based \abr{llm} backbones.}
In adapter-era systems, the language model remains the reasoning core and visual features are injected through a projector, resampler, or cross-attention bridge. GPT-4V~\cite{yang2023dawn}, LLaVA~\cite{liu2023visualinstructiontuning}, Qwen2-VL~\cite{wang2024qwen2vlenhancingvisionlanguagemodels}, DeepSeek-VL2~\cite{wu2024deepseekvl2mixtureofexpertsvisionlanguagemodels}, and Pixtral~\cite{agrawal2024pixtral12b} exemplify this recipe. The design is modular and efficient for instruction tuning, but it can leave vision as a secondary pathway rather than a fully shared representational space.

\noindent \textbf{Era 3a: native multimodal input, text output.}
The strongest 2025--2026 frontier families move beyond adapter-style fusion and instead pretrain a single multimodal trunk that directly consumes text, images, documents, audio, and video. In this regime, modalities are aligned earlier and more uniformly, while outputs remain primarily autoregressive text. Representative families include Gemini 3.6 Flash~\cite{google2026gemini36flash}, Claude Opus 4.6~\cite{anthropic2026claudeopus46}, GPT-5.6 Sol~\cite{openai2026gpt56}, Grok 4.5~\cite{xai2026grok45}, Qwen3.8-Max~\cite{qwen2026qwen38}, Kimi K3~\cite{kimi2026k3}, MiniMax M3~\cite{minimax2026m3}, and Gemma 4~\cite{google2026gemma4}. This design supports much longer context windows and makes multimodal perception central rather than auxiliary.

\noindent \textbf{Era 3b: unified multimodal input/output.}
Some families push further and ask the same trunk not only to understand multiple modalities, but also to generate them. Emu3~\cite{wang2024emu3nexttokenpredictionneed}, Janus-Pro~\cite{chen2025januspro}, and related unified models treat image, video, audio, and text as parts of one token or latent interface, typically with decoder heads such as VAE-, DiT-, or codec-based modules. The practical motivation is to collapse understanding and generation into one backbone rather than coupling separate specialist models.

\noindent \textbf{Era 4: world-action models.}
The newest step extends the multimodal trunk into a predictive control model. World-action systems such as Cosmos 3~\cite{nvidia2026cosmos3}, Kairos~\cite{kairos2026}, DreamX-World 1.0~\cite{dreamxworld2026}, and OmniDreams~\cite{omnidreams2026} treat action as another modality and optimize the model to predict future observations while selecting actions. In other words, the model becomes generator, perceiver, and policy in a single loop. Even when these systems are not yet the dominant general-purpose assistant architecture, they reveal where multimodal scaling is heading.

\subsection{VLM Alignments}
Alignment now affects not only safety and helpfulness, but also whether frontier \abr{vlm}s can sustain long reasoning traces over visual evidence. The broad recipe remains inherited from \abr{llm} alignment: supervised fine-tuning defines formats and tool use, preference optimization improves response quality and safety, and reinforcement learning increasingly targets hard multimodal reasoning skills.
Direct Preference Optimization (DPO)~\cite{rafailov2024directpreferenceoptimizationlanguage} and Proximal Policy Optimization (PPO)~\cite{schulman2017proximalpolicyoptimizationalgorithms} remain important foundations, but the strongest 2025--2026 trend is the adoption of GRPO-style rule-based reinforcement learning for visual reasoning, spatial reasoning, and multimodal generation.

\begin{table*}[htbp]
    \centering
    \small
    \renewcommand{\arraystretch}{1.15}
    \begin{tabular}{p{0.21\textwidth}p{0.08\textwidth}p{0.12\textwidth}p{0.20\textwidth}p{0.30\textwidth}}
        \hline
        \textbf{Method / Work} & \textbf{Year} & \textbf{Signal} & \textbf{Target setting} & \textbf{Why it mattered} \\
        \hline
        MM-RLHF~\cite{zhang2025mm} & 2025 & DPO & General multimodal preference alignment & Established large-scale human preference tuning for \abr{vlm}s and critique-based reward modeling \\
        MM-Eureka~\cite{meng2025mmeureka} & 2025 & RLOO~\cite{Kool2019Buy4R} & Visual reasoning & Showed that rule-based reinforcement learning can induce longer and stronger multimodal reasoning traces \\
        Vision-R1~\cite{huang2025visionr1incentivizingreasoningcapability} & 2025 & GRPO & Image reasoning & Brought DeepSeek-R1 style reasoning incentives into the \abr{vlm} setting \\
        R1-VL~\cite{zhang2025r1vllearningreasonmultimodal} & 2025 & Step-wise GRPO & Multimodal reasoning & Reinforced intermediate reasoning steps rather than only final answers \\
        S-GRPO~\cite{sgrpo2026} & 2026 & Supervised GRPO & Cold-start post-training & Combined ground-truth trajectories with GRPO to stabilize multimodal RL from weak initial policies \\
        Faithful GRPO~\cite{faithfulgrpo2026} & 2026 & Constrained GRPO & Visual-spatial reasoning & Used explicit consistency constraints to reduce shortcut reasoning and answer instability \\
        ZPPO~\cite{zppo2026} & 2026 & Prompt-zone policy optimization & Qwen-family \abr{vlm}s & Replaced gradient-heavy teacher distillation with prompt-level teacher guidance and improved generalization \\
        AlphaGRPO~\cite{alphagrpo2026} & 2026 & GRPO + decomposed rewards & Unified multimodal generation & Extended multimodal RL beyond understanding into generation and editing \\
        \hline
    \end{tabular}
    \caption{Representative post-training methods for modern \abr{vlm}s. The frontier moved from supervised instruction tuning and preference optimization toward multimodal reinforcement learning that can supervise reasoning, grounding, and generation with sparse or verifiable rewards.}
    \label{tab:rl_comparison}
\end{table*}

%
While RLHF succeeds on \abr{llm}s, \abr{vlm}s add at least two new alignment burdens: the model must reason over evidence that is only partially linguistic, and the reward signal must distinguish true visual grounding from plausible but unsupported verbal answers. When a model handles image inputs alongside text, it can reveal or infer sensitive details, over-trust its language prior, or misread crucial visual context. These concerns make multimodal alignment harder than merely porting text-only recipes.
Recent work therefore mixes several signals: human preference pairs~\cite{zhang2025mm}, rule-based correctness rewards~\cite{meng2025mmeureka, huang2025visionr1incentivizingreasoningcapability, zhang2025r1vllearningreasonmultimodal}, constrained rewards for spatial faithfulness~\cite{faithfulgrpo2026}, and teacher or verifier guidance that avoids the full cost of dense human labeling~\cite{zppo2026, alphagrpo2026, sgrpo2026}. The broader lesson is that post-training is now one of the main levers separating a capable multimodal base model from a practically useful reasoning model.
The key to RLHF is to collect human feedback and design reward models. In~\cite{zhang2025mm}, authors introduces a high-quality, human-annotated dataset with 120k preference comparison pairs to enhance the alignment of VLMs. It proposes a Critique-Based Reward Model that improves interpretability by generating critiques before assigning scores. By contrast, \cite{meng2025mmeureka, zhou2025r1} extends large-scale rule-based reinforcement learning to multimodal scenarios and reproduces key characteristics of text-based RL (like DeepSeek-R1 \cite{guo2025deepseek}) in visual contexts. Despite using only simple, sparse rewards (format and accuracy) and minimal data filtering, the authors achieve stable improvements in accuracy and response length. 

In addition to RLHF, Reinforcement Learning with Verifiable Rewards (RLVR) is also getting attention~\cite{liu2025visual}. RLVR bypasses the need for training a reward model by utilizing a direct verification function to evaluate correctness. This method streamlines the reward process while ensuring strong alignment with the task's correctness criteria.

\label{architecture}

\section{Benchmarks and Evaluation}
\label{sec:benchmarks}
The number of \abr{vlm} benchmarks has grown rapidly with the quick development of new \abr{vlm}s since 2022~\cite{chia2023instructevalholisticevaluationinstructiontuned, zhang2024visionlanguagemodelsvisiontasks}. 
Comprehensive benchmarking is important for evaluating model performance and ensuring robust training across different capabilities various aspects such as math reasoning, scene recognition, etc~\cite{MathVista, VQAv2}.
Modern \abr{vlm} benchmarks have moved far beyond basic VQA. The most recent frontier benchmarks test whether a model can track a scene over time, act through an embodied interface, maintain calibration on document extraction, or even serve as a reliable judge for long-horizon agents~\cite{fu2024mmesurveycomprehensivesurveyevaluation}.
In this section, we summarize representative benchmark categories rather than commit to a fixed count, because the benchmark frontier after 2025 changes monthly.
We first give a broad taxonomy of the tasks commonly used to evaluate \abr{vlm}s. We then highlight the 2025--2026 shift toward embodied, temporal, and judge-aware evaluation, summarize common metric choices, and discuss the limits of current automatic evaluation practice.
Our central observation is that benchmark diversity has increased faster than benchmark fidelity: many datasets broaden topics or modalities, but the hardest open problem is now whether the evaluation itself can faithfully measure grounded multimodal competence.


\noindent \textbf{Benchmark Categorization.} Benchmarks are designed with specific testing objectives, and we group them into a broad taxonomy of representative categories (Table~\ref{tab:benchmarks}).

\begin{table*}[!t]
\centering
\renewcommand{\arraystretch}{1.2}  
\scalebox{0.75}{
\begin{tabular}{p{0.3\textwidth}p{0.45\textwidth}p{0.45\textwidth}}
\hline
\textbf{Category} & \textbf{Description} & \textbf{Datasets} \\
\hline
Visual text understanding & Evaluates models' ability to extract and understand texts within visual components & TextVQA~\cite{TextVQA}, DocVQA~\cite{DocVQA} \\
Multilingual multimodal understanding & Evaluates VLMs on different languages on different tasks such as question answering and reasoning & MM-En/CN~\cite{MMBench}, CMMLU~\cite{CMMLU}, C-Eval~\cite{C-Eval}, MTVQA~\cite{MTVQA} \\
Visual math reasoning & Tests models' ability to solve math problems in image forms & MathVista~\cite{MathVista}, MathVision~\cite{MathVision}, MM-Vet~\cite{MM-Vet} \\
Optical Character Recognition (OCR) & Test models' ability to extract objects from visual inputs & MM-Vet~\cite{MM-Vet}, OCRBench~\cite{OCRBench}, MME~\cite{MME}, MMTBench~\cite{MMT-Bench} \\
Chart graphic understanding & Evaluates models' ability to interpret graphic-related data & infographic VQA~\cite{infographicVQA}, AI2D~\cite{AI2D}, ChartQA~\cite{ChartQA}, MMMU~\cite{MMMU} \\
Text-to-Image generation & Evaluates models' ability to generate images & MSCOCO~\cite{MSCOCO-30K}, GenEval~\cite{GenEval}, T2I-CompBench~\cite{T2I-CompBench}, DPG-Bench~\cite{DPG-Bench}, VQAScore~\cite{lin2024evaluatingtexttovisualgenerationimagetotext}, GenAI-Bench~\cite{li2024genaibenchevaluatingimprovingcompositional} \\
Hallucination & Evaluates whether models are likely to hallucinate on certain visual and textual inputs & HallusionBench~\cite{guan2024hallusionbench}, POPE~\cite{POPE}, CHAIR~\cite{rohrbach2018object}, M-HalDetect~\cite{gunjal2024detecting}, Hallu-Pi~\cite{ding2024hallu}, Halle-Switch~\cite{zhai2023halle}, BEAF~\cite{ye2025beaf}, AutoHallusion~\cite{wu2024autohallusionautomaticgenerationhallucination}, GAIVE~\cite{liu2023mitigating}, Hal-Eval~\cite{jiang2024hal}, AMBER~\cite{wang2023llm}  \\
Multimodal general intelligence & Evaluates models' ability on diverse domains of tasks & MMLU~\cite{MMLU}, MMMU~\cite{MMMU}, MMStar~\cite{MMStar}, M3GIA~\cite{M3GIA}, AGIEval~\cite{AGIEval} \\
Video understanding & Evaluates models' ability to understand videos (sequences of images) & EgoSchema~\cite{EgoSchema}, MLVU~\cite{MLVU}, MVBench~\cite{MVBench}, VideoMME~\cite{VideoMME}, MovieChat~\cite{song2024moviechatdensetokensparse}, Perception-Test~\cite{Perception-Test}, \\
Visual reasoning, understanding, recognition, and question answering & Evaluate VLMs' ability to recognize objects, answer questions, and reason through both visual and textual information & MMTBench~\cite{MMT-Bench}, GQA~\cite{GQA}, MM-En/CN~\cite{MMBench}, VCR~\cite{VCR}, VQAv2~\cite{VQAv2}, MM-Vet~\cite{MM-Vet}, MMU~\cite{MMBench}, SEEDBench~\cite{SEEDBench}, Real World QA~\cite{realWorldQA}, MMMU-Pro~\cite{MMMU-Pro}, DPG~\cite{DPG-Bench} , MSCOCO-30K~\cite{MSCOCO-30K}, MM-Vet~\cite{MM-Vet}, ST-VQA~\cite{st-vqa}, NaturalBench~\cite{li2024naturalbenchevaluatingvisionlanguagemodels}\\
Alignment with common sense and physics & Evaluate the alignment between the AIGC images and videos generated by \abr{vlm}s and the real world & VBench~\cite{huang2024vbench}, PhysBench~\cite{chow2025physbench}, VideoPhy~\cite{bansal2024videophy}, WISE~\cite{niu2025wiseworldknowledgeinformedsemantic}, VideoScore~\cite{he2024videoscore}, CRAVE~\cite{sun2025content}, WorldSimBench~\cite{qin2024worldsimbench}, WorldModelBench~\cite{li2025worldmodelbench}\\
GUI, web, and computer-use agents & Evaluates whether agents can ground interface elements and complete interactive tasks on screens, websites, phones, and operating systems & ScreenSpot~\cite{cheng2024seeclick}, AITW~\cite{rawles2023androidinthewild}, WebArena~\cite{zhou2023webarena}, VisualWebArena~\cite{koh2024visualwebarena}, OSWorld~\cite{xie2024osworld} \\
Robot benchmark and embodied control & Evaluate the embodied \abr{vlm}s' abilities online in rule-based simulators or offline datasets recording collected interactions & Habitat~\cite{habitat19iccv}, Gibson~\cite{xia2018gibson}, iGibson~\cite{li2021igibson}, Isaac Lab~\cite{mittal2023orbit}, CALVIN~\cite{mees2022calvinbenchmarklanguageconditionedpolicy}, VLMBench~\cite{zheng2022vlmbench}, GemBench~\cite{garcia2024generalizablevisionlanguageroboticmanipulation}, VIMA-Bench~\cite{jiang2022vima}, VirtualHome~\cite{puig2018virtualhome}, AI2-THOR~\cite{kolve2017ai2}, ProcTHOR~\cite{deitke2022procthor}, ThreeDWorld~\cite{gan2020threedworld} \\
Generative model, world model & Evaluate the embodied AI models' abilities with interactive models representing the environments & GAIA-1~\cite{hu2023gaia}, UniSim~\cite{yang2023learning}, LWM~\cite{liu2024worldmodelmillionlengthvideo}, Genesis~\cite{Genesis}, RoboGen~\cite{wang2023robogen}\\
\hline
\end{tabular}
}
\caption{Representative benchmark categories for evaluating \abr{vlm}s. Most historical benchmarks emphasize recognition, question answering, OCR, or generation quality, while newer work increasingly stresses temporal reasoning, embodied interaction, GUI/computer-use agents, and evaluation reliability in realistic settings such as web agents, robotics, and domain-specific decision making~\cite{liu2023vlpd, wu2023referring}.
}
\label{tab:benchmarks}
\vspace{-10pt}
\end{table*}

\begin{table*}[!t]
\centering
\small
\renewcommand{\arraystretch}{1.15}
\begin{tabular}{p{0.16\textwidth}p{0.10\textwidth}p{0.27\textwidth}p{0.37\textwidth}}
\hline
\textbf{Benchmark} & \textbf{Year} & \textbf{What it targets} & \textbf{What changed relative to older benchmarks} \\
\hline
HumanCLAW~\cite{humanclaw2026} & 2026 & Embodied multimodal decision making & Separates a \abr{vlm}'s decision quality from low-level motor execution, enabling attribution of failures to perception, planning, or actuation \\
GST-Bench~\cite{gstbench2026} & 2026 & Global spatial awareness from continuous video & Requires integrating a long traversal into a coherent scene model instead of solving local, single-view perception tasks \\
ChronoVision~\cite{chronovision2026} & 2026 & Multi-step temporal reasoning & Evaluates whether a model can reconstruct latent state over time instead of answering from one frame or a short clip \\
WorldExam~\cite{worldexam2026} & 2026 & Reactivity of world models & Distinguishes visual plausibility from behavioral correctness by asking whether a world model reacts properly rather than merely looks realistic \\
OSReward~\cite{osreward2026} & 2026 & Computer-use reward modeling & Evaluates the judge itself for long-horizon GUI agents, acknowledging that scoring agent trajectories is now a research problem \\
ExtractBench~\cite{extractbench2026} & 2026 & Enterprise document extraction & Moves document evaluation from free-form QA toward schema-grounded extraction with explicit structural targets \\
ConfBench~\cite{confbench2026} & 2026 & Confidence calibration for extraction & Measures whether the model knows when it is wrong, not only whether its extracted content matches the gold answer \\
RESPClinBench~\cite{respclinbench2026} & 2026 & Longitudinal clinical reasoning & Extends medical multimodal evaluation from single encounters to disease progression across time \\
\hline
\end{tabular}
\caption{Representative 2025--2026 benchmark directions. The strongest recent benchmarks are less concerned with whether a model can answer another isolated visual question and more concerned with whether it can act, remember, react, and judge reliably over longer horizons.}
\label{tab:recent_benchmarks}
\end{table*}

\begin{figure*}[htbp]
    \centering
    \begin{subfigure}[t]{0.45\textwidth}  
        \centering
        \includegraphics[width=\textwidth]{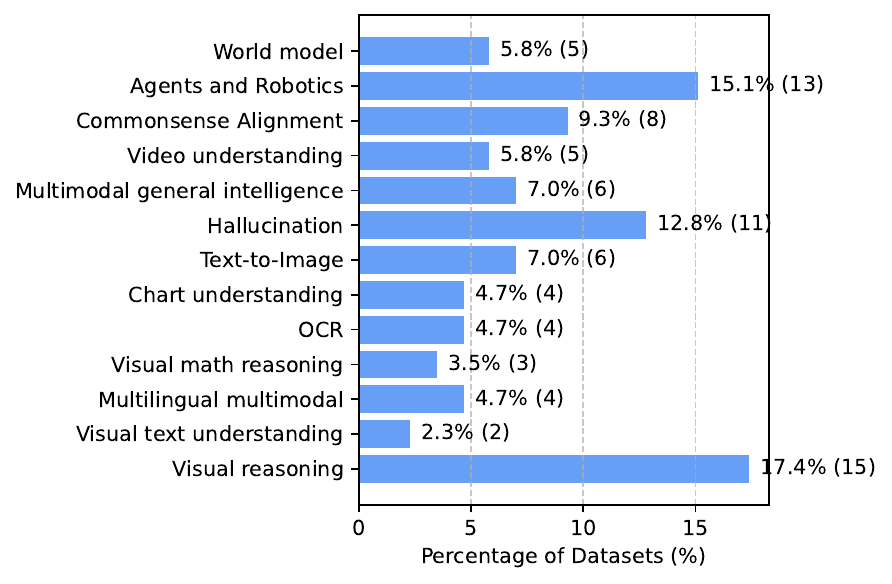}
        \caption{Most of our surveyed data tests VLMs' visual reasoning abilities.}
        \label{fig:bench_distribution}
    \end{subfigure}
    \hspace{0.05\textwidth}  
    \begin{subfigure}[t]{0.45\textwidth}  
        \centering
        \includegraphics[width=\textwidth]{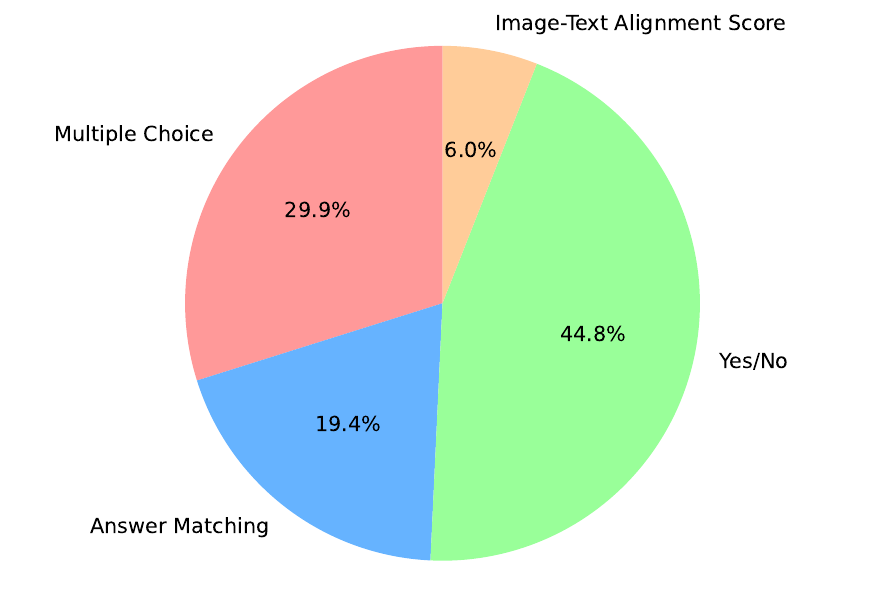}
        \caption{Majority of the benchmarks are designed in multiple choice or yes/no format for ease of evaluations.}
        \label{fig:eval_metric_2}
    \end{subfigure}
    \caption{Our surveyed (a) benchmark dataset categories and (b) common evaluation practices. Most existing benchmarks focus on Yes/No and multiple choice format for the ease of evaluation. However, multiple choice and Yes/No questions also have their limitations that VLMs/LLMs can blindly answer above random guessing probability without giving them the questions~\cite{balepur-etal-2024-artifacts}. Current scope of VLM benchmark and evaluation is broad but not comprehensive due to the challenges of reliability of answer matching.}
    \label{fig:eval_metrics}
\end{figure*}

\subsection{How Are Benchmark Data Collected}
Benchmark datasets are typically created using one of three common data collection pipelines: fully human-annotated datasets; partially human-annotated datasets scaled up with synthetic data generation and partially validated by humans; and partially human-annotated datasets scaled up with synthetic data and fully validated by humans.

\noindent \textbf{Fully human-annotated datasets} are created by having humans collect or generate adversarial or challenging test questions from diverse subjects and fields. 
For example, MMMU\cite{MMMU} has 50 college students from various disciplines to collect existing test questions from textbooks and lecture materials, often in multiple choice format. 
Another approach involves humans creating questions and having annotators provide answers to these questions. 
In VCR\cite{VCR}, Mechanical Turks are tasked with using contexts, detected objects, and images to write one to three questions about each image, along with reasonable answers and explanations.
Fully human annotated datasets are time-consuming and hard to scale up, which brings inspiration to automatic question generation with human validation.

\noindent \textbf{Synthetic question generation} has become a more popular part of benchmark generation pipeline on various disciplines such as chart understanding~\cite{ChartQA}, video understanding~\cite{EgoSchema} to quickly scale up dataset sizes.
Common practices include using human written examples as seed examples, giving a powerful \abr{llm} to generate more adversarial example questions and answers~\cite{SEEDBench}.
Often, the generation process is only involved with texts.
Chart and video data are often paired with visual content and captions, which are often used by authors as context to prompt \abr{llm}s to extract answers and generate questions~\cite{EgoSchema, MVBench}. 
However, \abr{llm}s are not always accurate and may produce unfaithful content or hallucinations without human supervision~\cite{xu2024hallucinationinevitableinnatelimitation, li2025largelanguagemodelsstruggle, gu2025largelanguagemodelseffective}. 
To address this, pipelines typically include automatic filters to remove low-quality outputs, followed by crowdworker validation of either randomly sampled or all generated examples~\cite{ChartQA, SEEDBench, EgoSchema}.
Automatic benchmark generation helps scale dataset size with reduced human effort. 
However, current automatic question-generation methods primarily rely on captions and textual contexts, which can lead to the creation of questions that are easy to answer without requiring significant visual reasoning~\cite{guan2024hallusionbench, mondal-etal-2024-scidoc2diagrammer}, which undermines the benchmark's primary goal—evaluating a \abr{vlm}'s ability to comprehend and reason about visual content.

\noindent \textbf{Interaction in the Simulator} is mainly targeted at \abr{vlm} benchmarks in robotics . 
It gathers data for training and evaluation by assessing the \abr{vlm}-powered agents online. 
As a data generation method stemming from reinforcement learning, such a data generation method is applicable for those scenarios that human-labeled datasets or synthetic datasets are hard and expensive to acquire, while the data construction follows some common rules like the physical law or some other common sense. With this rule-based data acquisition method, the outcome \abr{vlm}s are more robust to the deviation within the multimodal inputs. 
During recent years, many works focus on realistic environments for robotics~\cite{habitat19iccv, xia2018gibson, li2021igibson, mittal2023orbit, mees2022calvinbenchmarklanguageconditionedpolicy, zheng2022vlmbench, garcia2024generalizablevisionlanguageroboticmanipulation}, web agents~\cite{zhou2023webarena, koh2024visualwebarena}, mobile GUI control~\cite{rawles2023androidinthewild}, GUI grounding~\cite{cheng2024seeclick}, and open-ended computer use~\cite{xie2024osworld}. Nonetheless, benchmarks~\cite{habitat19iccv, xia2018gibson, li2021igibson} based on interaction traces or online execution are also widely used for \abr{vlm} agent training and evaluation. Notably, more efforts have been used for generative model~\cite{yang2023learning} or even world model~\cite{liu2024worldmodelmillionlengthvideo, hu2023gaia, Genesis} to replace the previous simulators or datasets in generating more practical and better-quality datasets for \abr{vlm}s.
Though simulators are widely used in training and evaluating the \abr{vlm}-power agents, the potential sim2real gap might exist when transplanting the terminal \abr{vlm} into real-world applications, \textit{i.e.} the \abr{vlm}-powered agents might not be able to handle some real-world situations. More efforts towards the mitigation of these issues are still expected in this direction.

\subsection{Evaluation Metrics}
\label{subsec:eval_metrics}
\begin{figure*}[htbp]
    \centering
    \includegraphics[width=0.9\textwidth]{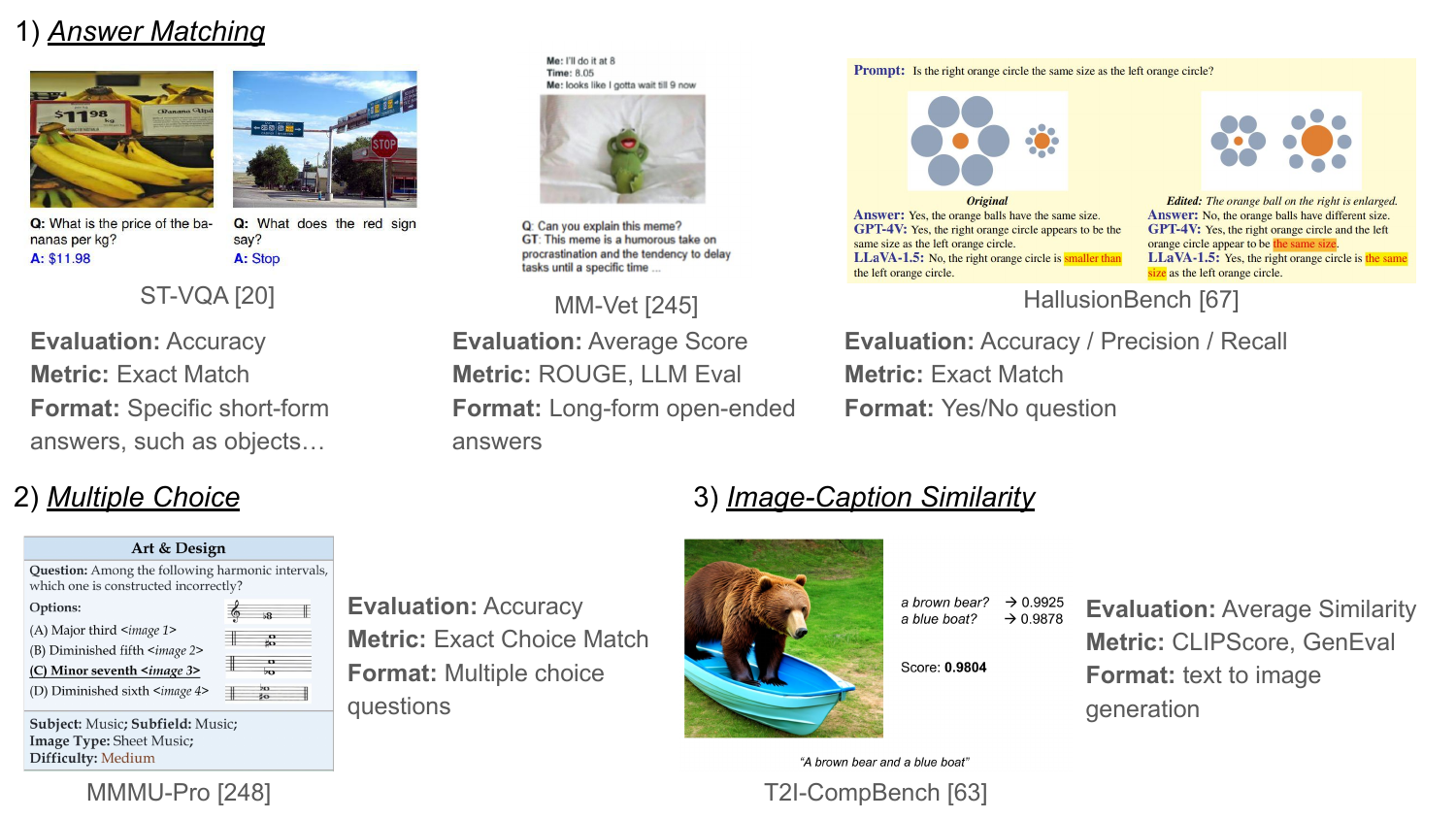}
    \caption{Common benchmark evaluation metrics restrict the formats of most benchmarks, which mostly evaluate whether a \abr{vlm} can generate a short-form answer that matches the correct answers.}
    \label{fig:eval_metric_demo}
\end{figure*}

Benchmarks are designed for evaluation, with metrics established during their creation. 
\abr{vlm} evaluation metrics are automatic to support repeated use at scale, and they often influence the question formats used in the benchmarks.
We show the common evaluation metrics used in our surveyed benchmarks (Figure~\ref{fig:eval_metric_2}, ~\ref{fig:eval_metric_demo}).

\noindent \textbf{Answer matching} is widely used for open-ended and closed-ended question types, where the answers are \textit{short-form entities, long-form answers, numbers, or yes/no}.
Generative \abr{vlm}s are more verbose than extractive \abr{llm}s and \abr{vlm}s, where they often generate verbose but correct answers~\cite{li-etal-2024-pedants}, containment exact match~\cite{izacard2021leveragingpassageretrievalgenerative} is a more practical version used more often in the evaluation, which includes removing articles and space of predicted answers and check whether the normalized predicted answer is contained in the normalized gold answer~\cite{lewis2021retrievalaugmentedgenerationknowledgeintensivenlp, chen2017readingwikipediaansweropendomain}.
However, exact match tends to have high recall, which often fails to account for semantic equivalence between the gold and predicted answers, frequently misjudging human-acceptable correct answers as incorrect~\cite{bulian2022tomayto, chen-etal-2019-evaluating, li-etal-2024-pedants} and becomes impossible for benchmarks that seek long-form answers~\cite{Xu2023ACE}.
Prior to the instruction following success of \abr{llm} period, standard token overlapping socres such as $F_1$, ROUGE~\cite{lin-2004-rouge}, BLEU~\cite{10.3115/1073083.1073135} to measure the similarity score between the gold and predicted answers, but start failing when generative models are generating more complex and diverse but correct answers~\cite{Xu2023ACE, chen-etal-2019-evaluating, li-etal-2024-pedants, bulian2022tomayto}.

Thus, some of the benchmarks like MM-Vet~\cite{MM-Vet} adopt \abr{llm}s to evaluate generated responses when the responses are long-form answers that requires semantic understanding to judge correctness.
\abr{llm} evaluations are shown to have the highest correlations to human evaluation, but they also face the struggles of producing consistent outputs with internal model updates or changing prompt instructions~\cite{manas2023improving, zhao2023large, kamalloo-etal-2023-evaluating}.
While no current answer-matching evaluation method is perfect, \textit{yes/no} questions are the easiest to evaluate compared to open-ended ones. 
As a result, most benchmarks rely on a multiple-choice format to assess \abr{vlm}s (Figure~\ref{fig:eval_metric_2}).

\noindent \textbf{Multiple Choice} format involves selecting an answer from a set of options, including distractors, for a given visual question~\cite{MMT-Bench, VCR, SEEDBench, realWorldQA}.
This format provides definitive answers and is among the easiest to evaluate, as it measures the percentage of questions a \abr{vlm} answers correctly. 
However, \abr{llm}s have demonstrated an unusual ability to select correct answers even without access to the actual questions~\cite{balepur-etal-2024-artifacts}. 
Since \abr{vlm}s incorporate an \abr{llm} component for generating responses (Section~\ref{sec:architecture}), further research is required to assess the robustness and reliability of current \abr{vlm} benchmarks.

\noindent \textbf{Judge reliability and calibration} are emerging as benchmark targets in their own right. Recent works such as OSReward~\cite{osreward2026}, ConfBench~\cite{confbench2026}, and WorldExam~\cite{worldexam2026} emphasize that for long-horizon agents, document extraction pipelines, and world models, verifying the output may be harder than generating it. This is a major departure from earlier benchmarks that assumed the scoring rule was fixed and unproblematic.

\noindent \textbf{Image/text similarity scores} are commonly used in image generation benchmarks like T2I-CompBench, GenEval~\cite{T2I-CompBench, GenEval} to evaluate the alignment between generated images and their corresponding textual descriptions. 
They often rely on measures such as CLIPScore~\cite{hessel2022clipscorereferencefreeevaluationmetric} for image-text alignment or ROUGE for caption matching to assess the semantic and lexical similarity between the outputs and the references.

In summary, \abr{vlm} benchmarks encompass a wide range of question types, fields of expertise, and tasks, with MMLU~\cite{MMLU} alone covering 57 distinct tasks. 
However, popular evaluations remain largely confined to simple answer matching or multiple choice formats, far from the broader general intelligence of the Turing test~\cite{turing1950computing}.

\label{benchmarks}

\section{Applications}
\label{sec:application}
Modern \abr{vlm} applications are no longer limited to captioning or short-form \abr{vqa}. As frontier families gained stronger grounding, longer context, and better tool use, they became practical in closed-loop settings such as screen control, web navigation, document workflows, robotics, and domain-specific decision support. These deployments also reveal the gap between benchmark performance and real-world reliability more clearly than static offline evaluation.

\subsection{Computer-Use, GUI, and Web Agents}
One of the clearest new application fronts is computer use: agents that read screenshots or rendered interfaces, ground language to visual elements, and emit actions such as clicks, taps, text entry, or browser navigation. ScreenAI~\cite{baechler2024screenaivisionlanguagemodelui} focuses on understanding user interfaces and infographics, CogAgent~\cite{hong2024cogagent} improves high-resolution GUI grounding, and ScreenAgent~\cite{niu2024screenagentvisionlanguagemodeldriven} turns screenshot perception into a planning-acting-reflecting loop for real screen interaction. Octopus~\cite{yang2024octopusembodiedvisionlanguageprogrammer}, ShowUI~\cite{qinghong2024showui}, and WebVoyager~\cite{he2024webvoyager} further illustrate the move from passive screen understanding to end-to-end action on phones, desktops, and real websites.

This category matters because it stresses several capabilities at once: fine-grained OCR, element grounding, long-horizon planning, tool invocation, and recovery from execution errors. In other words, GUI and web agents turn \abr{vlm}s from visual assistants into interactive operators, which is why recent benchmarks increasingly evaluate computer-use and reward-model quality rather than only answer accuracy.

\subsection{Document Intelligence and Accessibility}
Another rapidly growing application layer is document-centered reasoning. Interfaces, reports, dashboards, forms, and charts are often more important in practice than natural images, and they require models to combine OCR, layout understanding, numerical extraction, and grounded language generation. ScreenAI~\cite{baechler2024screenaivisionlanguagemodelui} is representative of this shift because it targets UI and infographic understanding directly, while ChartLLaMA~\cite{han2023chartllamamultimodalllmchart} shows how specialized post-training can improve chart interpretation where generic models often hallucinate values or axes.

These same capabilities support accessibility-oriented applications. SciDoc2Diagrammer~\cite{mondal-etal-2024-scidoc2diagrammer} converts scientific documents into diagrams, and recent systems for immersive or web-based accessibility use \abr{vlm}s to generate contextual visual descriptions for users with low vision~\cite{oliveira2024improving}. The practical importance of this area is high because errors are easy for end users to notice: a model that misreads a table cell, a chart label, or a form field is not merely imprecise, but directly unhelpful.

\subsection{Generative Media and Design}
\abr{vlm}s are also increasingly used as creative tools rather than only analytical ones. In lighter-weight consumer settings, systems such as Supermeme.ai~\cite{supermeme2024} use multimodal understanding and generation to produce memes and social content. At the high end, video-generation systems such as Movie Gen~\cite{polyak2024movie} point toward multimodal assistants that can storyboard, edit, and synthesize richer media assets from mixed text-image-video inputs.

This application category is especially relevant to the transition from Era 3a to Era 3b systems described earlier. Once a model family can both understand and generate visual content through a shared multimodal trunk, the boundary between analysis, editing, and creation becomes much thinner.

\subsection{Robotics and Autonomous Systems}
In physical-world settings, \abr{vlm}s act as perception-and-reasoning modules for manipulation, navigation, and driving. VIMA~\cite{jiang2022vima} and Instruct2Act~\cite{huang2023instruct2act} show how language-conditioned visual reasoning can support manipulation, while the RT series~\cite{o2023open, brohan2022rt, brohan2023rt} demonstrates the broader vision-language-action direction. For navigation, systems such as ZSON~\cite{majumdar2022zson} and VLFM~\cite{yokoyama2024vlfm} use grounded multimodal representations to guide exploration and object-goal search. In autonomous driving, DriveVLM~\cite{tian2024drivevlmconvergenceautonomousdriving} and related systems use multimodal reasoning for scene understanding, planning, and long-tail case analysis.

These applications are among the most demanding because they convert model outputs into physical or safety-critical behavior. As a result, they expose weaknesses in calibration, latency, sim-to-real transfer, and action reliability more starkly than chatbot-style deployments do.

\subsection{High-Stakes Domain Assistants}
Some of the most valuable near-term uses of \abr{vlm}s are expert copilots in domains where images are semantically dense but humans remain in the loop. In healthcare, VisionUnite~\cite{li2024visionunite} and clinician-evaluated radiology studies~\cite{yildirim2024multimodal} illustrate how multimodal models can support diagnosis, explanation, and patient communication. In education, recent work uses \abr{vlm}s for diagram and mathematics reasoning~\cite{wu2024analyzing} and even for simulating students to help teachers practice instruction~\cite{ma2024students}. Related directions appear in agriculture~\cite{zhu2024harnessing}, \abr{ai} for science~\cite{cao2024promoting}, and climate applications~\cite{chen2024vision}.

Across these domains, the pattern is consistent: \abr{vlm}s are moving from passive describers to interactive copilots that must ground perception, reason over domain context, and communicate uncertainty appropriately. That shift is one reason why post-training, reward modeling, and embodied or calibration-aware evaluation have become central to modern \abr{vlm} development.

\section{Challenges of \abr{vlms}}
\label{sec: challenges}

This section examines key challenges in \abr{vlm} research, including hallucination, safety, fairness, alignment, efficiency in training and fine-tuning, and data scarcity. Despite recent advancements, understanding these limitations is crucial to mitigating risks and ensuring ethical, reliable deployment, particularly for marginalized users.

\subsection{Hallucination}

Hallucination in \abr{vlm}s refers to referencing nonexistent objects in images~\cite{rohrbach2018object}. Despite benchmark-setting performance, hallucination is still a pervasive issue in \abr{vlm}s, especially in visual-text tasks. Researchers have proposed datasets and metrics to quantify hallucination, with early efforts tending to require human annotation.
CHAIR~\cite{rohrbach2018object} quantifies hallucination in image captioning using per-instance and per-sentence metrics.
POPE~\cite{POPE} assesses hallucination with Yes-No questions on object existence.
M-HalDetect~\cite{gunjal2024detecting} provides 16K fine-grained VQA samples for training \abr{vlm}s to detect and prevent hallucinations.

Subsequent research investigated hallucination in finer detail. 
Halle-Switch~\cite{zhai2023halle} evaluates hallucination based on data amount, quality, and granularity, balancing contextual and parametric knowledge.
Hallu-Pi~\cite{ding2024hallu} provides 1,260 images with detailed annotations to detect perturbation-induced hallucinations.
BEAF~\cite{ye2025beaf} examines before-and-after image changes, introducing new metrics: true understanding, ignorance, stubbornness, and indecision.
HallusionBench~\cite{guan2024hallusionbench} tests VLM visual reasoning with dependent, unanswerable questions across diverse topics and formats.
AutoHallusion~\cite{wu2024autohallusionautomaticgenerationhallucination} automates hallucination benchmark generation, probing \abr{vlm} language modules for context-based hallucination examples.

The advent of more sophisticated \abr{llm}s has also assisted the development of larger benchmark datasets in this area. 
GAIVE~\cite{liu2023mitigating} uses GPT-4 to generate 400K samples across 16 vision-language tasks, covering hallucinations like nonexistent object and knowledge manipulation.
Hal-Eval~\cite{jiang2024hal} constructs 2M image-caption pairs, leveraging GPT-4 for fine-grained hallucination induction.
AMBER~\cite{wang2023llm} is an \abr{llm}-free multi-dimensional benchmark designed for generative and discriminative tasks, annotating four types of hallucination.

\subsection{Safety}
Given \abr{vlm}s' versatility, safeguarding against unethical use is crucial. Jailbreaking~\cite{jin2024jailbreakzoo} allows malicious circumvention of ethical boundaries, posing risks in robotics and other downstream tasks~\cite{yue2024safe, duan2024aha, wu2024highlightingsafetyconcernsdeploying, robey2024jailbreaking, karnik2024embodied}.
SafeBench \cite{ying2024safebench} introduces harmful queries across 23 risk scenarios using an \abr{llm}-based jury deliberation framework.
MM-SafetyBench \cite{liu2024mm} evaluates \abr{vlm} safety with image-text query pairs in unsafe contexts.
%
%
JailbreakV \cite{luo2024jailbreakv} introduces 28K malicious image-based queries, testing attack transferability across models.
SHIELD \cite{shi2024shield} evaluates face spoofing and forgery detection using True-False queries in zero- and few-shot settings.
HADES \cite{li2024images} exploits gradient updates and adversarial methods to conceal and amplify harmful content, breaking multimodal alignment.
imgJP \cite{niu2024jailbreaking} bypasses refusal guardrails using images instead of prompts, demonstrating high transferability across  \abr{vlm}s.


\subsection{Fairness}

Extensive literature has explored inequities in \abr{llm}s and \abr{vlm}s \cite{bai2024hallucination, gallegos2024bias}. Like unimodal \abr{llm}s, \abr{vlm}s show disparate performance, particularly affecting marginalized groups \cite{adewumi2024fairness, hundt2022robots, azeem2024llm}. MMBias \cite{janghorbani2023multi} presents a human-annotated image dataset targeting bias in religion, nationality, disability, and sexual orientation. FMBench \cite{wu2024fmbench} proposes a benchmark using medical images to assess gender, skin tone, and age bias. Harvard-FairVL \cite{luo2024fairclip} shows CLIP and BLIP2 favor Asian, Male, and Non-Hispanic groups. FairmedFM \cite{jinfairmedfm} integrates 17 datasets to evaluate fairness in medical tasks. CulturalVQA \cite{nayak2024benchmarking} (2,378 image-question pairs) shows better performance for North American cultures and worse performance for African and Islamic ones.


\subsection{Alignment}
\textbf{Multi-modality Alignment.} 
The alignment issue in multi-modal models arises from contextual deviation between modalities, leading to hallucinations~\cite{wang2024mitigating}. Many efforts to mitigate this include leveraging VLM reasoning for self-reflection~\cite{wang2024enhancing} or designing projectors to bridge modalities.
SIMA~\cite{wang2024enhancing} improves L-VLM alignment via self-critique and vision metrics.
SAIL~\cite{zhang2024assessinglearningalignmentunimodal} aligns pretrained unimodal models for better multimodal learning.
Ex-MCR~\cite{wang2023extending} enables paired-data-free semantic alignment using contrastive representation.
OneLLM~\cite{wang2023extending} unifies eight modalities to language through a unified encoder and progressive multimodal alignment.
SeeTRUE~\cite{yarom2023you} benchmarks text-image alignment, proposing VQA-based and end-to-end classification methods for better misalignment detection and ranking.

\noindent \textbf{Commonsense and Physics Understanding.} 
The L\abr{vlm}s used for AI-generated content (AIGC) images and videos, sometimes known as World Models~\cite{ha2018world}, like SORA~\cite{brooks2024video} and Veo2~\cite{veo2}, attract much attention throughout the community. However, these L\abr{vlm}s face challenges in commonsense alignment and physics adherence. Many recent benchmarks and evaluation models aim to address these issues.
VBench~\cite{huang2024vbench} evaluates video generative models across structured quality dimensions.
PhysBench~\cite{chow2025physbench} and VideoPhy~\cite{bansal2024videophy} assess VLMs and text-to-video models on physical understanding and commonsense adherence.
WISE~\cite{niu2025wiseworldknowledgeinformedsemantic} introduces WiScore for T2I knowledge-image alignment.
CRAVE~\cite{sun2025content} focuses on AIGC video quality assessment, aligning textual prompts with video dynamics.
VideoScore~\cite{he2024videoscore} tracks model progress using VideoFeedback, a large-scale human-annotated dataset.
WorldSimBench~\cite{qin2024worldsimbench} and WorldModelBench~\cite{li2025worldmodelbench} evaluate World Simulators for video-action consistency and decision-making applications.
GPT4Motion~\cite{lv2024gpt4motion} integrates \abr{llm}s, physics engines, and diffusion models for physics-aware text-to-video synthesis.
Despite these efforts, key challenges remain in advanced video evaluation and bridging the gap between AIGC and real-world fidelity.

\noindent \textbf{Training Efficiency.}
Efficient training and alignment of \abr{vlm}s remain a very heated research topic due to their high cost and difficulty in training. Recent studies explore the impact of different pre-training settings over modules~\cite{lin2024vila} or supervision~\cite{wang2021simvlm} on the ultimate performance of \abr{vlm}s. However, many applications require specialized rather than multi-task capabilities. Low-Rank Adaptation (LoRa)~\cite{hu2021lora, dettmers2023qloraefficientfinetuningquantized} enables efficient fine-tuning with fewer parameters. RLHF~\cite{bai2022training, lee2023rlaif} integrates human or model feedback for alignment. Rule-based RL, requiring multiple input generations, increases computational costs, limiting its use to small \abr{vlm}s~\cite{peng2025lmmr1empowering3blmms}. Alternative RL methods (PPO, DPO) reduce computation but demand extensive human annotation to trade for computation resources~\cite{meng2025mmeureka, zhang2025mm}.

 \vspace{-0.1 cm}
\subsection{Data Scarcity}
The abilities and reliabilities of \abr{vlm}s are highly depending on the availability and diversity of the training datasets. However, the massive scale of current advanced \abr{vlm}s and the scarcity of high-quality training datasets add up to the difficulty in continuously improving the performance of the future \abr{vlm}s. One potential method to mitigate this issue is to use self-supervised learning (SSL)~\cite{mu2022slip} that learns the representation automatically from the unlabelled dataset. Another major direction is to use the synthetic data generated by following some rules~\cite{balazadeh2024syntheticvisiontrainingvisionlanguage} or utilizing some third-party tools~\cite{sharifzadeh2024synth2boostingvisuallanguagemodels}. In \abr{vlm}, specifically designed for physical world-related purposes, like robotics~\cite{tang2024kalie} or web agents~\cite{chae2024webagentsworldmodels}, another option is to gather datasets from the interactions with the physical simulators or world model~\cite{wang2023robogen, katara2024gen2sim, wu2023daydreamer, Genesis}, or learning from videos with human demonstrations~\cite{liang2024dreamitate, zhou2025you}. Though a lot of efforts have been made in all three directions, more insights are still expected into the breakthrough of the mass-scale training for L\abr{vlm}s and the alternatives to the internet-scale data, given Ilya Sutskever's quote that “Pre-training as we know it will unquestionably end.”

\vspace{-0.1 cm}
\section{Conclusion}
Developments of  \abr{vlm}s and \abr{llm}s are happening at a breakneck pace with more sophisticated applications and use cases being introduced in quick succession. This paper aims to capture the most notable architectures, trends, applications along with prominent challenges in this area. We hope that our survey provides a solid general overview for practitioners as a road map for future works. 
\label{challenges}

\section{Limitation}

Given the rapid growth of \abr{vlm} research, our survey is intentionally selective rather than exhaustive. We focus on representative frontier families, benchmark directions, and post-training methods that clarify the field's main architectural and evaluation shifts through 2026. However, the pace of releases remains faster than any static manuscript can track. New models, benchmarks, and alignment recipes will continue to appear after publication, so the companion repository and website remain the better medium for near-real-time updates.
\label{limitation}


{
    \small
    \bibliographystyle{ieeenat_fullname}
    \bibliography{main}
}


\end{document}